# 4chan & 8chan embeddings

Pierre Voué, Tom De Smedt, Guy De Pauw

**AUTHORS**

- **Pierre Voué** (pierre@textgain.com) has a master's degree in Artificial Intelligence and works as a data scientist at Textgain, studying Salafi-jihadism and right-wing extremism.
- **Tom De Smedt** has a PhD in Arts and is CTO at Textgain, focusing on Natural Language Processing. He was awarded the Research Prize of the Auschwitz Foundation in 2019.
- **Guy De Pauw** has a PhD in Linguistics and is CEO of Textgain. He has over 20 years of experience in Natural Language Processing and Machine Learning.

**ABSTRACT**

We have collected over 30M messages from the publicly available /pol/ message boards on 4chan and 8chan, and compiled them into a model of toxic language use. The trained word embeddings (±0.4GB) are released for free and may be useful for further study on toxic discourse or to boost hate speech detection systems: https://textgain.com/8chan.

## 1 INTRODUCTION

4chan (4chan.org, 2003) and 8chan (8ch.net, 2013) are largely unregulated online message boards that have been linked to white supremacy (Shultz, 2019), right-wing extremism (Davey, Saltman & Birdwell, 2018), hate crimes (Powell & Henry, 2017), harassment (Thibault, 2016), misogyny (Jane, 2016) and disinformation (Marwick & Lewis, 2017). To illustrate this, in August 2014, an online harassment campaign targeting women in the video game industry (Gamergate) was manufactured on 4chan, before spreading to Twitter and Reddit and drawing worldwide attention (Chess & Shaw, 2015). Similary, in November 2016, a conspiracy theory falsely implicating high-ranking officials of the US Democratic Party in human trafficking and child sexual abuse (Pizzagate) originated on 4chan and 8chan, before spreading to other social media platforms (Tuters, Jokubauskaitė & Bach, 2018).

The /pol/ boards (*pol* meaning *Politically Incorrect*) are particularly notorious and have been linked to violent extremism in real-life, such as the Christchurch mosque shootings (March 2019),[1] the Poway synagogue shooting (April 2019) and the El Paso Walmart shooting (August 2019).[2] The unregulated boards essentially serve as a relay for the shooter's worldviews, by disseminating, discussing and also idolizing their manifestos. In August 2019, "cesspool of hate" 8chan was taken offline by their service provider,[3] but has recently resurfaced as 8kun.top, now without a /pol/ board.

The toxic discourse that 4chan and 8chan users engage in is also known as *shitposting*,[4,5] referring to large amounts of offensive content meant to upset readers, often in combination with *edgy memes*,[6] low-quality satirical images on the border of illegal. A popular and effective strategy in this regard is dehumanization: comparing humans to animals, diseases and objects, and/or stripping them of their humanity by ridiculing their cognitive capabilities for example (see Jaki & De Smedt, 2018). Popular targets include *cucks* (weak or effeminate men), *femoids* (woman in general perceived as subhuman), *social justice warriors* (civil rights activists), *snowflakes* (left-wing do-gooders), *kikes* (Jews), *dindu nuffins* (black people), *goat fuckers*, and so on.

Online hate speech is on the rise,[7] and a focus point of the European Commission, IT companies and academics.[8] From a language technology perspective, 4chan and 8chan constitute a vast dataset of toxic language, where new terminology originates before spreading to mainstream platforms.

## 2 METHODS AND MATERIALS

Using the Python programming language, we collected a dataset of over 30 million /pol/ messages posted between November 2013 and August 2019. The corpus is available upon motivated request. About 90% of the data is from 4chan, which is archived externally by archive.4plebs.org. However, because both 4chan and 8chan regularly delete older content, and 8chan is not archived, we were only able to collect messages from 8chan from July 2019 onward, constituting about 10% of the corpus.

---

[1] https://www.bellingcat.com/news/americas/2019/04/28/ignore-the-poway-synagogue-shooters-manifesto-pay-attention-to-8chans-pol-board

[2] https://www.vox.com/recode/2019/5/3/18527214/8chan-walmart-el-paso-shooting-cloudflare-white-nationalism

[3] https://blog.cloudflare.com/terminating-service-for-8chan

[4] https://www.adl.org/resources/reports/gab-and-8chan-home-to-terrorist-plots-hiding-in-plain-sight

[5] https://www.urbandictionary.com/define.php?term=shitposting

[6] https://www.urbandictionary.com/define.php?term=edgy+meme

[7] https://www.europol.europa.eu/activities-services/main-reports/terrorism-situation-and-trend-report-2019-te-sat

[8] https://ec.europa.eu/commission/news/countering-illegal-hate-speech-online-2019-feb-04_en

Word embedding (Mikolov et al., 2013) is a fairly new approach in Machine Learning, a subfield of AI that focuses on automated “learning by example”. In broad strokes, word embedding consists of examining the context of each word in a large collection of texts, since words derive meaning from the company they keep. For each word, we can count the words that precede and follow it, i.e., its context. Then, knowing the context, we can discover unknown words that frequently occur in a similar context. For example, the words *nigger* and *faggot* often occur in similar sentences on 4chan and 8chan, whereas they seldom occur on other social media.

The model was created as follows. First, every /pol/ message was standardized by removing URLs and platform-specific metadata, and by converting it to lowercase. Then, to train the distributional semantics model, we used the Continuous Bag-of-Words algorithm (CBOW) from Gensim (Řehůřek & Sojka, 2011) with window size 5 (words before and after), retaining stop words (*the*, *and*, *maybe*, etc.). The output vectors have an average size of 150, which can be thought of as “how condensed” a word’s meaning is. Finally, we used negative sampling to improve the model’s learning speed.

## 3 RESULTS

The resulting model can be downloaded at https://textgain.com/8chan, in binary and in raw format. There is a range of word embedding models in industry and academia, typically aiming to capture the semantics of a given language or in-group jargon. We have created a model that captures the toxicity of the 4chan and 8chan community. It is freely available and can be easily integrated in Deep Learning systems. To our knowledge, it is the only resource of its kind.

## 4 DISCUSSION

We can now verify that *cuck* (for example) is indeed used as a pejorative reference to weak and effeminate men. In Listing 1, a simple Python script lists the 10 most similar words (0.0-1.0) on 4chan and 8chan as: *cuckold* (0.8), *faggot* (0.8), *soyboy* (0.7), *fag* (0.7), *loser* (0.7), *weakling* (0.7), and so on:

**Listing 1**. Example Python code for querying the embeddings, using Grasp.py.

```
from grasp import Embedding # https://github.com/textgain/grasp

from io import open

f = open('48chan_embeddings.txt')

e = Embedding.load(f, format='txt')

for p, w in e.similar('cuck', n=10):
    print(w, round(p, 2))

for p, w in e.similar('nigger', n=10):
    print(w, round(p, 2))
```

Querying the embeddings for the word *nigger* exposes related words like *nignog*, *monkey*, *darkie*, *jigaboo*, *coon*, i.e., which are used in a similar context (see also Figure 1). In this way, the model can be leveraged to populate a list of well-known and more obscure pejorative words.

As shown in Listing 2, it is not difficult to expand the list iteratively by also searching on other social media, for example using the Twitter API. In this case, we find a tweet stating “It’s called acting you nignog hoodbooger,” containing the unknown word *hoodbooger*, which we can add to our list.

**Listing 2**. Example Python code for iterative keyword expansion, using Grasp.py.

```
from grasp import twitter

for _, w in e.similar('nigger'):
    for tweet in twitter.search(w):
        for w in tweet.text.split():
            if w not in e:
                print(w)
```

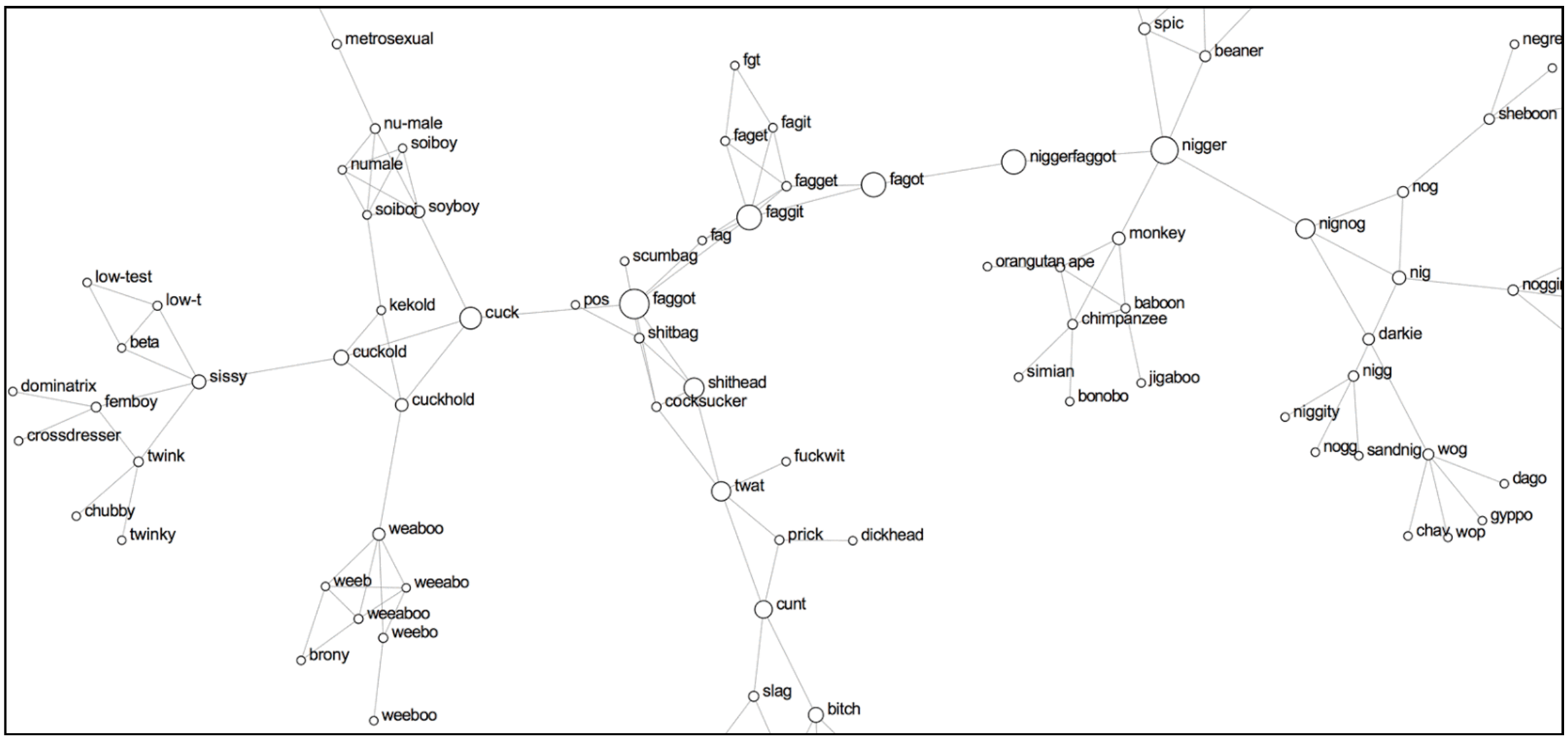

**Figure 1.** Related words in the model, represented as a network.

**ACKNOWLEDGEMENTS**

The initial prototype of the model was trained during Google Summer of Code 2019 (GSoC).